\documentclass[letterpaper]{article} 
\usepackage{aaai23}  
\usepackage{times}  
\usepackage{helvet}  
\usepackage{courier}  
\usepackage[hyphens]{url}  
\usepackage{graphicx} 
\usepackage{amsmath}

\usepackage{natbib}  
\usepackage{caption} 
\usepackage{algorithm}
\usepackage{algorithmic}

\usepackage{newfloat}
\usepackage{listings}
\DeclareCaptionStyle{ruled}{labelfont=normalfont,labelsep=colon,strut=off} 
\floatstyle{ruled}
\newfloat{listing}{tb}{lst}{}
\floatname{listing}{Listing}
\title{Causal Inference Based Single-branch Ensemble Trees For Uplift Modeling
}
\author{
    Fanglan Zheng,
    Menghan Wang\equalcontrib,
    Kun Li\equalcontrib,
    Jiang Tian,
    Xiaojia Xiang
}
\affiliations{
    \textsuperscript{\rm}Everbright Technology Co. LTD, Beijing, China\\

}

\usepackage{bibentry}

\begin{document}

\maketitle

\begin{abstract}
In this manuscript (ms), we propose causal inference based single-branch ensemble trees for uplift modeling, namely CIET. Different from standard classification methods for predictive probability modeling, CIET aims to achieve the change in the predictive probability of outcome caused by an action or a treatment. According to our CIET, two partition criteria are specifically designed to maximize the difference in outcome distribution between the treatment and control groups. Next, a novel single-branch tree is built by taking a top-down node partition approach, and the remaining samples are censored since they are not covered by the upper node partition logic. Repeating the tree-building process on the censored data, single-branch ensemble trees with a set of inference rules are thus formed.
Moreover, CIET is experimentally demonstrated to outperform previous approaches for uplift modeling in terms of both area under uplift curve (AUUC) and Qini coefficient significantly. At present, CIET has already been applied to online personal loans in a national financial holdings group in China. CIET will also be of use to analysts applying machine learning techniques to causal inference in broader business domains such as web advertising, medicine and economics.
\end{abstract}

\section{Introduction}
Uplift modeling involves a set of methods for estimating the expected causal impact of taking an action or a treatment at an individual or subgroup level, which could lead to an increase in their conversion probability \cite{Zhao2019UpliftMF}.
Typically for financial services and commercial companies looking to provide additional value-added services and products to their customers, marketers may be interested in evaluating the effectiveness of numerous marketing techniques, such as sending promotional coupons.
With the change in customer conversion possibilities, marketers are able to efficiently target prospects.
More than marketing campaigns, uplift modeling can be applied to a variety of real-world scenarios related to personalization, such as online advertising, insurance, or healthcare, where patients with varying levels of response to a new drug are identified, including the discovery of adverse effects on specific subgroups \cite{Jaskowski2012UpliftMF}.

In essence, uplift modeling is a problem that combines causal inference and machine learning. For the former, it is mutually exclusive to estimate the change between two outcomes for the same individual. To overcome this counterfactual problem, samples are randomly assigned to a treatment group (receiving online advertisement or marketing campaign) and a control group (not receiving online advertisement nor marketing campaign). For the latter, the task is to train a model that predicts the difference in the probability of belonging to a given class on the two groups.
At present, two major categories of estimation techniques have been proposed in the literature, namely meta-learners and tailored methods \cite{Zhang2022AUS}. The first includes the Two-Model approach \cite{Radcliffe2007UsingCG}, the X-learner \cite{Knzel2017MetalearnersFE} and the transformed outcome methods \cite{Athey2015MachineLM} which extend classical machine learning techniques. The second refers to direct uplift modeling such as uplift trees \cite{5693998} and various neural network based methods \cite{ Louizos2017CausalEI,Yoon2018GANITEEO}, which modify the existing machine learning algorithms to estimate treatment effects. 
Also, uplift trees can be extended to more general ensemble tree models, such as causal forests \cite{doi:10.1080/01621459.2017.1319839,10.1214/18-AOS1709}, at the cost of losing true interpretability.

In order to take advantage of decision trees and the ensemble approach, we propose causal inference based single-branch ensemble trees for uplift modeling (CIET) with two completely different partition criteria that directly maximizes the difference between outcome distributions of the treatment and control groups. When building a single-branch tree, we employ lift gain and lift gain ratio as loss functions or partition criteria for node splitting in a recursive manner. 
Since our proposed splitting criteria are highly related to the incremental impact, the performance of CIET is thus expected to be reflected in the uplift estimation. Meanwhile, 
the splitting logic of all nodes along the path from root to leaf is combined to form a single rule to ensure the interpretability of CIET. Moreover, the dataset not covered by the rule is then censored and the above tree-building process continue to repeat on the censored data. 
Due to this divide and conquer learning strategy, the dependencies between the formed rules can be effectively avoided. It leads to the formation of single-branch ensemble trees and a set of decorrelated inference rules.  

Note that our CIET is essentially different from decision trees for uplift modeling and causal forests. There are three major differences: (1) single-branch tree $vs$ standard binary tree; (2) lift gain and its ratio as loss function or splitting criterion $vs$ Kullback-Leibler divergence and squared Euclidean distance; and (3) decorrelated inference rules $vs$ correlated inference rules or even no inference rules. 
It is demonstrated through empirical experiments that CIET can achieve better uplift estimation compared with the existing models. Extensive experimental results on synthetic data and the public credit card data show the success of CIET in uplift modeling. We also train an ensemble model and evaluate its performance on a large real-world online loan application dataset from a national financial holdings group in China. As expected, the corresponding results show a significant improvement in the evaluation metrics in terms of both AUUC and Qini coefficient.


The rest of this ms is organized as follows. First, causal inference based single-branch ensemble trees for uplift modeling is introduced. Next, full details of our experimental results on synthetic data, credit card data and real-world online loan application data are given. 
It is demonstrated that CIET performs well in estimating causal effects compared to decision trees for uplift modeling. Finally, conclusions are presented.

\section{Causal Inference Based Single-Branch Ensemble Trees (CIET) for Uplift Modeling}\label{sec:Alg}
This section consists of three parts. We first present two splitting criteria, single-branch ensemble approach and pruning strategy specially designed for the uplift estimation problem. Evaluation metrics for uplift modeling are then discoursed. Three key algorithms of CIET are further described in detail.

\subsection{Splitting Criteria, Single-Branch Ensemble Method and Pruning Strategy}
Two distinguishing characteristics of CIET are splitting criteria for tree generation and the single-branch ensemble method, respectively.

As for splitting criteria in estimating uplift, it is motivated by our expectation to achieve the maximum difference between the distributions of the treatment and control groups. 
Given a class-labeled dataset with $N$ samples, $N^{T}$ and $N^{C}$ are sample size of the treatment and control groups (recall that $N = N^{T} + N^{C}$, $T$ and $C$ represent the treatment and control groups).
Formally, in the case of a balanced randomized experiment, the estimator of the difference in sample average outcomes between the two groups is given by:
\begin{align}
    \label{tau} \tau = (P^{T} - P^{C})(N^{T} + N^{C})
\end{align}
where $P^{T}$ and $P^{C}$ are the probability distribution of the outcome for the two groups. Motivated by Eq. (\ref{tau}), the divergence measures for uplift modeling we propose are lift gain and its ratio, namely LG and LGR for short. The corresponding mathematical forms of LG and LGR can be thus expressed as
\begin{align}
    \label{LG} LG &=(P_{R}^{T} - P_{R}^{C})N_{R} - \tau_{0} = \tau_{R} - \tau_{0}\\
    \label{LGR} LGR  &=\frac{(P_{R}^{T} - P_{R}^{C})}{(P_{0}^{T} - P_{0}^{C})} \propto (P_{R}^{T} - P_{R}^{C}) = \frac{\tau_{R}}{N_{R}}
\end{align}
where $P_{0}^{T}$ and $P_{0}^{C}$ are the initial probability distribution of the outcome for the two groups, $\tau_{0} = (P_{0}^{T} - P_{0}^{C})N_{R}$. And $N_{R}$ and $Y_{R}$ for a node logic $R$ represent coverage and correction classification,
while $P_{R}^{T}$ and $P_{R}^{C}$ are the corresponding probability distribution of the outcome for both groups, respectively. Evidently, both Eq. (\ref{LG}) and Eq. (\ref{LGR}) represent the estimator for uplift, which are proposed as two criteria in our ms. Compared to the standard binary tree with left and right branches, only one branch is created after each node splitting in this ms. It is characterized by the fact that both LG and LGR are calculated using the single-branch observations that present following a node split. Accordingly, subscript $k$ indicating binary branches doesn't exist in the above equation. Furthermore, the second term of LG makes every node partition better than randomization, while LGR has the identical advantages to information gain ratio.

The proposed splitting criterion for a test attribute A is then defined for any divergence measure $D$ as 
\begin{equation}\label{CRITERION}
    \Delta = D(P^T(Y):P^C(Y)|A) - D(P^T(Y):P^C(Y))  
\end{equation}
where $D(P^T(Y):P^C(Y)|A)$ is the conditional divergence measure. Apparently, $\Delta$ is the incremental gain in divergence following a node splitting. Substituting for $D$ the $LG$ and $LGR$, 
we obtain our proposed splitting criteria $\Delta_{LG}$ and $\Delta_{LGR}$. 
The intuition behind these splitting criteria is as follows: we want to build a single-branch tree such that the distribution divergence between the treatment and control groups before and after splitting an attribute differ as much as possible. 
Thus, an attribute with the highest $\Delta$ is chosen as the best splitting one. In order to achieve it, we need to calculate and find the best splitting point for each attribute. In particular, an attribute is sorted in descending order by value when it is numerical. 
For categorical attributes, some encoding methods are adopted for numerical type conversion. The average of each pair of adjacent values in an attribute with $n$ value, forms $n-1$ splitting points or values. 
As for this attribute, the point of the highest $\Delta$ can be seen as the best partition one. 
Furthermore, the best splitting attribute with the highest $\Delta$ can be achieved by traversing all attributes. 
As for the best splitting attribute, the instances are thus divided into two subsets at the best splitting point. One feeds into a single-branch node, while the other is censored. 
Note that the top–down, recursive partition will continue unless there is no attribute that explains the incremental estimation with statistical significance. Also, histogram-based method can be employed to select the best splitting for each feature, which can reduce the time complexity effectively.

Due to noise and outliers in the dataset, a node may merely represent these abnormal points, resulting in model overfitting. Pruning can often effectively deal with this problem. That is, using statistics to cut off unreliable branches. Since none of the pruning methods is essentially better than others, we use a relatively simple pre-pruning strategy. If $\Delta$ gain is less than a certain threshold, node partition would stop. Thus, a smaller and simpler tree is constructed after pruning. Naturally, decision-makers prefer less complex inference rules, since they are considered to be more comprehensible and robust from business perspective.

\subsection{Evaluation Metrics for Uplift Modeling}
As noted above, it is impossible to observe both the control and treatment outcomes for an individual, which makes it difficult to find measure of loss for each observation.
It leads that uplift evaluation should differ drastically from the traditional machine learning model evaluation. 
That is, improving the predictive accuracy of the outcome 
does not necessarily indicate that the models will have better performance in identifying targets with higher uplift. In practice, most of the uplift literature resort to aggregated
measures such as uplift bins or curves. Two key metrics involved are area under uplift curve (AUUC) and Qini coefficient \cite{Gutierrez2016CausalIA}, respectively. In order to define AUUC, binned uplift predictions are sorted from largest to smallest. For each $t$, the cumulative sum of the observations statistic is formulated as below,
\begin{equation}\label{AUUC}
    f(t) = (\frac{Y_t^{T}}{N_t^{T}} - \frac{Y_t^{C}}{N_t^{C}})(N_t^{T} + N_t^{C}) 
\end{equation}
where the $t$ subscript implies that the quantity is calculated on the first or top $t$ observations.
The higher this value, the better the uplift model. The continuity of the uplift curves makes it 
possible to calculate AUUC, i.e. area under the real uplift curve, which can be used to evaluate and compare different models. As for Qini coefficient, it represents a natural generalization of 
Gini coefficient to uplift modeling. Qini curve is introduced with the following equation,
\begin{equation}\label{QINI_Curve}
   g(t) = {Y_t^{T}} - \frac{Y_t^{C}N_t^{T}}{N_t^{C}}
\end{equation}
There is an obvious parallelism with the uplift curve since $f(t)=g(t)(N_t^{T}+N_t^{C})/N_t^{T}$.
The difference between the area under the actual Qini curve and that under the diagonal corresponding to random targeting can be obtained. 
It is further normalized by the area between the random and the optimal targeting curves, which is defined as Qini coefficient.

\subsection{Algorithm Modules}
The following representation of three algorithms includes: selecting the best split for each feature using the splitting criteria described above, learning a top-down induced single-branch tree and forming ensemble trees with each resulting tree progressively.

Algorithm \ref{single1:algorithm} depicts how to find the best split of a single feature $F$ on a given dataset $D[group\_key, feature,$ $ target]$ using a histogram-based method with the proposed two splitting criteria. Gain\_left and Gain\_right are the uplift gains for the child nodes after each node partition. If the maximum value of Gain\_left is greater than that of Gain\_right, the right branch is censored and vice versa. Thus, the best split with its corresponding splitting logic, threshold and uplift gain is found, which is denoted by Best$\_$Direction, Best$\_$Threshold and Best$\_\Delta$. Besides, there are several thresholds to be initialized before training a CIET model, including minimum number of samples at a inner node $min\_samples$, minimum recall $min\_recall$ and minimum uplift gain required for splitting $min\_\Delta$. The top-down process would continue only when the restrictions are satisfied.

\begin{algorithm}[htbp]
\caption{Selecting the Best Split for One Feature}
\label{single1:algorithm}
\textbf{Input}: D, the given class-labeled dataset, including the group key (treatment/control);\\
\textbf{Parameter}: feature $F$, min$\_$samples, min$\_$recall, min$\_\Delta$\\
\textbf{Output}: the best split that maximizes the lift gain or lift gain ratio on a feature

\begin{algorithmic}[1] 
\STATE \textbf{Set} Best$\_$Value = 0, Best$\_$Direction = "", Best$\_\Delta$ = 0, Best$\_$Threshold = None
\STATE calculate $Y^T$, $Y^C$, $N^T$, $N^C$ on D
\STATE For each feature value $v$ , calculate the values of $Y_{F\leq v}^T$, $Y_{F\leq v}^C$, $N_{F\leq v}^T$, $N_{F\leq v}^C$, and then the Gain\_left($v$) and Gain\_right($v$) with LG \eqref{LG} or LGR \eqref{LGR}.
\STATE set the Gain\_left($v$) and Gain\_right($v$) to their minimum value on the $v$, whose split does not satisfy the restrictions on number of samples/recall rate/divergence measure gain.  
\STATE $v_1$ = argmax(Gain\_left($v$)), $v_2$ = argmax(Gain\_right($v$))
\IF {max(Gain\_left) $\geq$ max(Gain\_right)}
\STATE Best$\_$Value = max(Gain\_left)
\STATE Best$\_$Direction = "$\leq$"
\STATE Best$\_$Threshold = $v_1$
\ELSE
\STATE Best$\_$Value = max(Gain\_right)
\STATE Best$\_$Direction = "$>$"
\STATE Best$\_$Threshold = $v_2$
\ENDIF
\STATE \textbf{return} Best$\_$Value, Best$\_$Direction, Best$\_$Threshold
\end{algorithmic} 
\end{algorithm}

Algorithm \ref{single:algorithm} presents a typical algorithmic framework for top–down induction of a single-branch uplift tree, which is built in a recursive manner using a greedy depth-first strategy. 
The parameter $max\_depth$ represents the depth of the tree and $cost$ indicates the threshold of LG or LGR.
As the tree grows deep, more instances are censored since they are not covered by the node partition logic of each layer. 
As a result, each child node subdivides the original dataset hierarchically into a smaller subset until the stopping criterion is satisfied. Tracing the splitting logics on the path from the root to leaf nodes in the tree, an "IF-THEN" inference rule is thus extracted. 

Finally, adopting a divide-and-conquer strategy, the above tree-building process is repeated on the censored samples to form ensemble trees, resulting in the formation of a set of inference rules as shown in Algorithm \ref{multi:algorithm}.

\begin{algorithm}[htbp]
\caption{Learning An "IF-THEN" Uplift Rule of A Single-branch Tree}
\label{single:algorithm}
\textbf{Input}: D, the given class-labeled dataset, including the group key (treatment/control);\\
\textbf{Parameter}: max$\_$depth, cost, min$\_$samples, min$\_$recall, min$\_\Delta$\\
\textbf{Output}: an "IF-THEN" uplift rule

\begin{algorithmic}[1] 
\STATE \textbf{Set} Rule$\_$Single = [], Max$\_$Gain = 0.0
\STATE \textbf{Set} Add\_Rule = \textbf{True}
\WHILE{depth $\leq$ max$\_$depth \textbf{and} Add\_Rule}
\IF {the treatment group or control group in D is empty}
\STATE break
\ENDIF
\STATE \textbf{Set}  Keep = \{ \}, Best$\_$Split = \{ \}
\STATE depth $\leftarrow $ depth + 1
\STATE Add\_Rule = \textbf{False}
\FOR{feature in features}
\STATE Keep[feature] =  Best\_Split\_for\_One\_Feature(D, feature,  min$\_$samples, min$\_$recall, min$\_\Delta$) (Algorithm \ref{single1:algorithm})
\ENDFOR
\FOR{feature in Keep}
\IF {feature's best gain $>$ Max\_Gain + cost}
\STATE Max$\_$Gain = feature's best gain
\STATE \textbf{Add} Keep[feature] \textbf{to} Best$\_$Split
\STATE Add\_Rule = \textbf{True}
\ELSE
\STATE continue
\ENDIF
\ENDFOR
\STATE \textbf{Add} Best$\_$Split \textbf{to} Rule$\_$Single
\STATE D $\leftarrow $ D $\setminus$ \{
Samples covered by Rule$\_$Single\} 
\ENDWHILE
\STATE \textbf{return} Rule$\_$Single
\end{algorithmic} 
\end{algorithm}

\begin{algorithm}[htbp]
\caption{Learning A Set of "IF-THEN" Uplift Rules}
\label{multi:algorithm}
\textbf{Input}: D, the given class-labeled dataset, including the group key (treatment/control);\\
\textbf{Parameter}: max$\_$depth, rule$\_$count, cost, min$\_$samples, min$\_$recall, min$\_\Delta$\\
\textbf{Output}: a set of "IF-THEN" uplift rules

\begin{algorithmic}[1] 
\STATE \textbf{Set} Rule$\_$Set = \{\}, number = 0
\WHILE{number $\leq$ rule$\_$count}
\STATE rule = Single$\_$Uplift$\_$Rule(D, cost, max$\_$depth, min$\_$samples, min$\_$recall, min$\_\Delta$) (Algorithm \ref{single:algorithm})\\
\STATE \textbf{Add} rule \textbf{to} Rule$\_$Set  
\STATE D $\leftarrow $ D $\setminus$ dataset covered by rule 
\STATE number $\leftarrow$ number + 1
\ENDWHILE
\STATE \textbf{return} Rule$\_$Set
\end{algorithmic} 
\end{algorithm}

\begin{figure*}[htbp]
\centering
\includegraphics[width=0.95\textwidth]{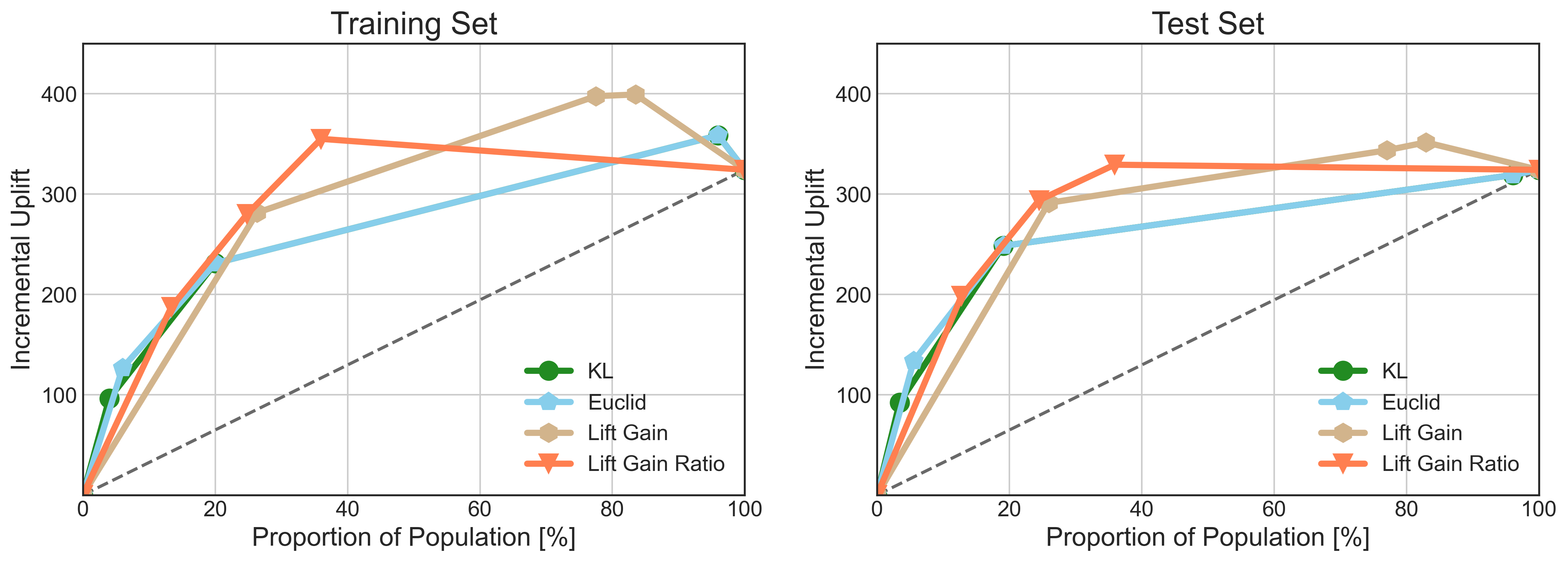} 
\caption{The uplift curves of four analyzed classifiers with four different colors for the synthetic dataset, while the dashed line corresponding to random targeting.}
\label{fig:AUUC_SSD} 
\end{figure*}

\section{Experiments}\label{sec:er}
In this section, the effectiveness of our CIET is evaluated on synthetic and real-world business datasets. Since CIET fundamentally stems from tree-based approaches, we implement it and compare it with uplift decision trees based on squared Euclidean distance and Kullback-Leibler divergence \cite{5693998}, which are referred as baselines.

\begin{table*}[ht]
\centering
\begin{tabular}{llll}
\hline
rule number  &"1"  &"2"  &"3"\\
\hline
node logic     &x9$\_$uplift $\leq$ 0.17                                  &x3$\_$informative $>$ -1.04                               &x6$\_$informative $>$ 0.95 \\
node logic     &x10$\_$uplift $\leq$ 2.59                                 &x1$\_$informative $\leq$ 2.71                             &x1$\_$informative $\leq$ 1.58\\
node logic     &null                                                      &x2$\_$informative $\leq$ 1.28                             &x9$\_$uplift $\leq$ 2.09     \\
$N_{before}$     &3000     &2210     &675     \\
$N_{before}^T$   &1500     &1117     &348     \\
$N_{before}^C$   &1500     &1093     &327     \\
$N_{rule}$        &790     &1535     &180     \\
$N_{rule}^T$      &383      &769      &76     \\
$N_{rule}^C$      &407      &766     &104     \\
net gain       &195.99    &87.14   &37.66     \\
$recall_{treatment}$  &36.62$\%$ &70.42$\%$ &42.11$\%$ \\
$recall_{control}$    &28.00$\%$ &63.70$\%$ &44.90$\%$ \\
\hline
\end{tabular}
\caption{A set of inference rules found by CIET and their corresponding statistical indicators with $criterion\_type = $"LG", $rule\_count = 3$ and $max\_depth = 3$.}
\label{tab:RS_CIET}
\end{table*}

\subsection{Experiments on Synthetic Data}\label{subsec: SD}
\textbf{Dataset} We can test the methodology with numerical simulations. That is, generating synthetic datasets with known causal and non-causal relationships between the outcome, action (treatment/control) and some confounding variables. More specifically, both the outcome and the action/treatment variables are binary. A synthetic dataset is generated with the $make\_uplift\_classification$ function in "Causal ML" package, based on the algorithm in \cite{Guyon2003DesignOE}.
There are 3,000 instances for the treatment and control groups, with response rates of 0.6 and 0.5, respectively. The input consist of 11 features in three categories. 8 of them are used for base classification, which are composed of 6 informative and 2 irrelevant variables. 2 positive uplift variables are created to testify positive treatment effect. The remaining one is a mix variable, which is defined as a linear superposition of a randomly selected informative classification variable and a randomly selected positive uplift variable.

\textbf{Parameters and Results} There are four hyper-parameters in CIET: $criterion\_type$, $max\_depth$ and $rule\_count$. $criterion\_type$ includes two options, LG and LGR. 
More precisely, two main factors of business complexity and difficulty in online deployment, determine parameter assignment.
Due to the requirement of model generalization and its interpretability, $max\_depth$ is set to 3. That is, the business logics of a single inference rule are always less than or equal to 3. And, $rule\_count$ is given a value of 3, indicating that a set of no more than three rules is defined to model the causal effect of a treatment on the outcome. Meanwhile, the default values for $min\_samples$, $min\_recall$, $cost$ and $min\_\Delta$ are 50, 0.1, 0.01 and 0, respectively.

\begin{table}[htbp]
\centering
\begin{tabular}{lllll}
\hline
dataset    &KL  &Euclid  &LG  &LGR\\
\hline
training   &0.187  &0.189  &0.239  &0.235 \\
test       &0.176  &0.178  &0.210  &0.225 \\
\hline
\end{tabular}
\caption{Qini coefficients of four analyzed classifiers on the training and test sets of the synthetic data.}
\label{tab:QC_SSD}
\end{table}

The stratified sampling method is used to divide the synthetic dataset into training and test sets in a ratio of fifty percent to fifty percent. Figure \ref{fig:AUUC_SSD} shows the uplift curves of the four analyzed classifiers. The AUUC of CIET with LG and LGR are 294 and 292 on the training set, which are significantly greater than 266 and 265 for the decision trees for uplift modeling with KL divergence and squared Euclidean distance. 
At the 36th percentile of the population, the cumulative profit increase reach 303 and 354 for LG and LGR, resulting in a growth rate of more than 18$\%$ and 37$\%$ compared to baselines. Besides, AUUC shows little variation in the training and test datasets, indicating that the stability of CIET is also excellent. According to Table \ref{tab:QC_SSD}, Qini coefficients of CIET are also obviously greater, with an increase of more than 24.5$\%$ and 17.8$\%$. Furthermore, all three rules are determined by uplift and informative variables as expected, which can be seen from Table \ref{tab:RS_CIET}.


\subsection{Experiments on Credit Card Data}

\textbf{Dataset} We use the publicly available dataset $Credit$ $Approval$ from the UCI repository as one of the real-world examples, which contains 690 credit card applications. All the 15 attributes and the outcome are encoded as nonsense symbols, where $A7 \neq v$ 
is applied as the condition for dividing the dataset into treatment and control groups. 
There are 291 and 399 observations in the two groups with response rates of 0.47 and 0.42, respectively.
Attributes with more than 25$\%$ difference in distribution between the two groups should be removed before any experiments are performed. This leads that 12 attributes are left as input variables. For further preprocessing, categorical features are binarized through one-hot encoding.

\begin{figure}[htbp]
\centering
\includegraphics[width=0.95\linewidth]{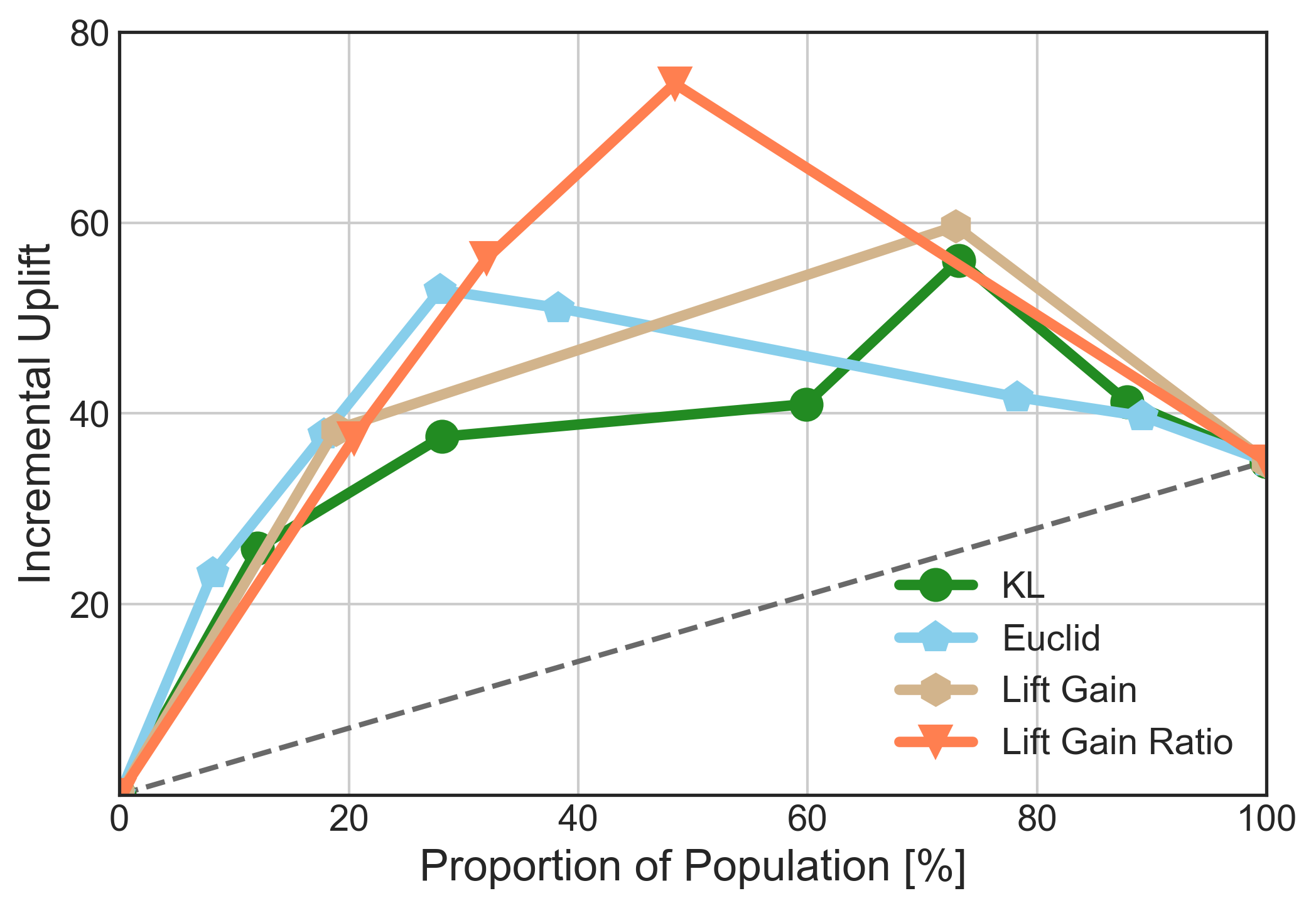} 
\caption{The uplift curves of four analyzed classifiers with four different colors for the $Credit\ Approval$ dataset, while the dashed curve corresponding to the random targeting.}
\label{fig:FIG_CAD} 
\end{figure}

\begin{table}[htbp]
\centering
\begin{tabular}{lllll}
\hline
metrics    &KL  &Euclid  &LG  &LGR\\
\hline
AUUC   &37.337  &40.887  &42.893  &48.222 \\
Qini   &0.201  &0.236  &0.257  &0.310 \\
\hline
\end{tabular}
\caption{Model performance of four analyzed classifiers on the $Credit\ Approval$ dataset.}
\label{tab:CAD}
\end{table}


\textbf{Parameters and Results} Based on the business decision-making perspective, the initial parameters are also the same as above.


\begin{figure*}[htbp]
\centering
\includegraphics[width=0.95\textwidth]{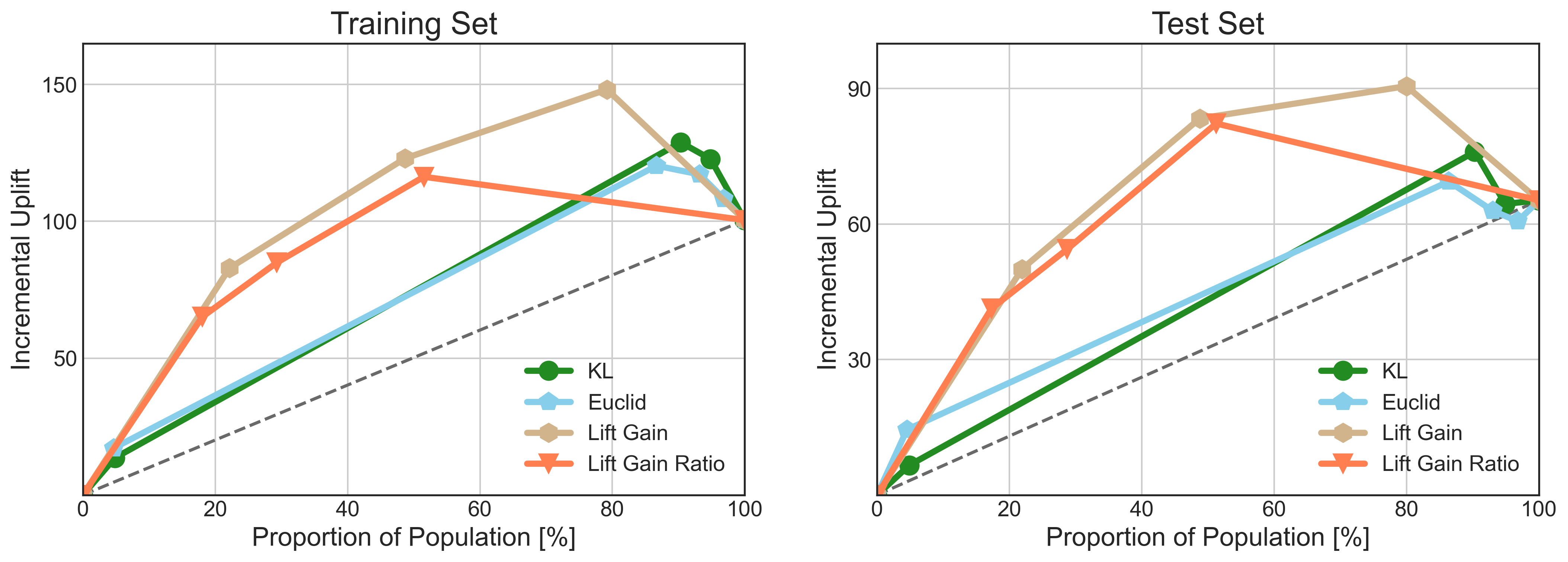} 
\caption{The uplift curves of four analyzed classifiers with four different colors for the real world online loan application dataset, while the dashed line corresponding to random targeting.}
\label{fig:AUUC_ROP} 
\end{figure*}

In order to avoid the distribution bias caused by the division on such small sample size dataset, there is no need to divide $Credit\ Approval$ into training and test parts. Figure \ref{fig:FIG_CAD} shows the uplift curves for the four analyzed classifiers, from which we can see that CIET is able to obtain higher AUUC and Qini coefficients. As shown in Table \ref{tab:CAD}, the former increases from 37$\sim$40 at baselines to 42$\sim$48 at CIET approximately, while the latter also improves significantly from 0.20$\sim$0.23 to 0.25$\sim$0.31. Especially when LGR serves as the splitting criterion, the cumulative profit has a distinguished peak of 74.5, while only 48.4$\%$ of the samples are covered.

\subsection{Experiments on Online Loan Application Data}

\textbf{Dataset} We further extend our CIET to precision marketing for new customer application. A telephone marketing campaign is designed to promote customers to apply for personal credit loans at a national financial holdings group in China via its official mobile app. The target is 1/0, indicating whether a customer would submit an application or not. 
The data contains 53,629 individuals, consisting of a treated group of 32,984 (receiving marketing calls) and a control group of 20,645 (not receiving marketing calls). These two groups have 300 and 124 credit loan applications with response rates of 0.9\% and 0.6\%, which are typical values in real world marketing practice. There are 24 variables in all, which are characterized as credit card-related information, loan history, customer demographics et al.

\begin{table}[htbp]
\centering
\begin{tabular}{lllll}
\hline
dataset    &KL  &Euclid  &LG  &LGR\\
\hline
training   &0.173  &0.168  &0.414  &0.302 \\
test       &0.108  &0.124  &0.385  &0.319 \\
\hline
\end{tabular}
\caption{Qini coefficients of four analyzed classifiers on the training and test sets of the real world online loan application data.}
\label{tab:QC_ROP}
\end{table}



\textbf{Parameters and Results} All parameters are the same as in the above experiments. The dataset is first divided into training and test sets in a ratio of sixty percent to forty percent. The response rates are consistent across two sets for two groups. Figure \ref{fig:AUUC_ROP} diplays the results graphically. As for the training dataset, CIET based on LG and LGR reach AUUC of about 104 and 89, while the decision trees based on KL divergence and squared Euclidean distance are 73 and 72. It can be seen that CIET achieves a significant improvement compared to baselines on this real-world dataset even with a very low response rate. Moreover, as can be seen in Table \ref{tab:QC_ROP}, Qini coefficient based on our approaches increases to 0.30$\sim$0.41 from 0.16$\sim$0.17 
on the training dataset. Meanwhile, Qini coefficient changes little when crossing to test dataset, indicating a better stability. Consequently, classifier with our CIET for precision marketing is effectively improved as well as stabilized in terms of AUUC and Qini coefficient. At present, CIET has already been applied to personal credit telemarketing.

\section{Conclusion}\label{sec:con}
In this ms, we propose new methods for constructing causal inference based single-branch ensemble trees for uplift estimation, CIET for short. Our methods provide two partition criteria for node splitting and strategy for generating ensemble trees. The corresponding outputs are uplift gain between the two outcomes and a set of interpretable inference rules, respectively. Compared with the classical decision tree for uplift modeling, CIET can not only be able to avoid dependencies among inference rules, but also improve the model performance in terms of AUUC and Qini coefficient. It would be widely applicable to any randomized controlled trial, such as medical studies and precision marketing. 

\bibliography{aaai23}

\end{document}